\documentclass[11pt,a4paper]{article}
\usepackage[hyperref]{emnlp2020}
\usepackage{times}
\usepackage{latexsym}

\usepackage{epsfig}
\usepackage{graphicx}
\usepackage{amsmath}
\usepackage{amssymb}
\usepackage{array, multirow}
\usepackage{amsfonts}
\usepackage{mathtools}
\usepackage{rotating} 
\usepackage{rotfloat}

\usepackage{microtype}

\aclfinalcopy 
\def\aclpaperid{1009} 

\title{BiST: Bi-directional Spatio-Temporal Reasoning \\for Video-Grounded Dialogues}

\author{Hung Le$^{\dag}{^\S}$\thanks{\hspace{0.1cm} This work was mostly done when Hung Le was an intern at Salesforce Research Asia, Singapore.}
, Doyen Sahoo$^{\ddag}$, Nancy F. Chen$^{\S}$, Steven C.H. Hoi$^{\dag\ddag}$ \\
  $^{\dag}$ Singapore Management University \\
  \texttt{hungle.2018@smu.edu.sg} \\
  $^{\ddag}$ Salesforce Research Asia\\
  \texttt{\{dsahoo,shoi\}@salesforce.com} \\ 
  $^\S$Institute for Infocomm Research, A*STAR \\ 
  \texttt{nfychen@i2r.a-star.edu.sg}}

\date{}

\begin{document}
\maketitle
\begin{abstract}
Video-grounded dialogues are very challenging due to (i) the complexity of videos which contain both spatial and temporal variations, and (ii) the complexity of user utterances which query different segments and/or different objects in videos over multiple dialogue turns. However, existing approaches to video-grounded dialogues often focus on superficial temporal-level visual cues, but neglect more fine-grained spatial signals from videos. To address this drawback, we propose Bi-directional Spatio-Temporal Learning (BiST), a vision-language neural framework for high-resolution queries in videos based on textual cues. Specifically, our approach not only exploits both spatial and temporal-level information, but also learns dynamic information diffusion between the two feature spaces through spatial-to-temporal and temporal-to-spatial reasoning. The bidirectional strategy aims to tackle the evolving semantics of user queries in the dialogue setting. The retrieved visual cues are used as contextual information to construct relevant responses to the users. Our empirical results and comprehensive qualitative analysis show that BiST achieves competitive performance and generates reasonable responses on a large-scale AVSD benchmark. We also adapt our BiST models to the Video QA setting, and substantially outperform prior approaches on the TGIF-QA benchmark. 
\end{abstract}

\section{Introduction}
\begin{figure}[h]
	\centering
	\resizebox{1.0\columnwidth}{!} {
	\includegraphics{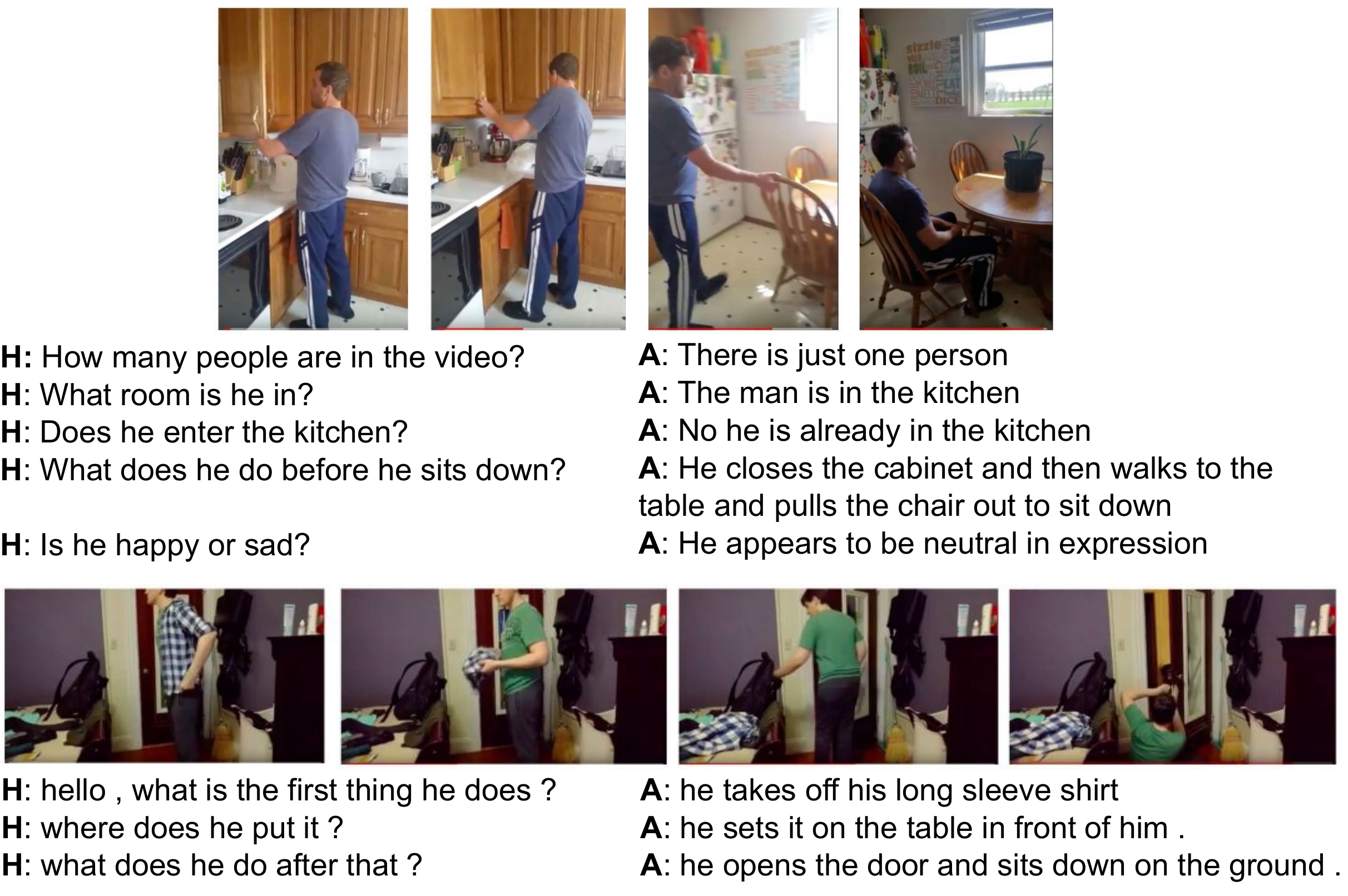}
	}
	\caption{Examples of video-grounded dialogues from the benchmark datasets of Audio-Visual Scene Aware Dialogues (AVSD) challenge \cite{alamri2018audio, alamri2019audiovisual}.
	\textbf{H}: human,  \textbf{A}: the dialogue agent.}
	\label{fig:samples}
\end{figure}

A video-grounded dialogue agent aims to converse with humans not only based on signals from natural language but also from other modalities such as sound and vision of the input video. 
Recent efforts \cite{alamri2018audio, sanabria2019cmu, alamri2019audiovisual} consider video-grounded dialogues as an extension of video Question-Answering (QA) \cite{MovieQA, jang2017tgif, lei-etal-2018-tvqa} whereby the agent answers questions from humans over multiple turns rather than a single turn (See Figure \ref{fig:samples}).
This is a very complex task as the dialogue agent needs to possess not only strong language understanding to generate natural responses but also sophisticated reasoning over video information, including the related objects, their positions and motions, etc.
Compared to image-based NLP tasks such as image QA and captioning \cite{antol2015vqa, xu2015show, goyal2017making}, video-grounded dialogues are more challenging as the feature representation of a video involves both spatial and temporal dimensions. 
Ideally, a dialogue agent has to process information of both dimensions to address the two major questions: ``where to look" (spatial reasoning) and ``when to look" (temporal reasoning) in the video. 

However, current approaches in video-grounded dialogues \cite{hori2019avsd, le-etal-2019-multimodal, sanabria2019cmu} often overlook spatial features and assume each spatial region is equally important to the current task (each spatial region is assigned with a uniform weight). Such approach is appropriate for cases where the video involves just few objects and spatial positions can be treated similarly. 
However, in many scenarios (e.g. examples in Figure \ref{fig:samples}), each video frame often contains multiple distinct objects and not all of them are relevant to the given question.

Related tasks to video-grounded dialogues are video QA and video captioning. 
Previous efforts in these research areas such as \cite{jang2017tgif, aafaq2019spatio} explicitly consider both spatial and temporal features of input video.
These models learn to summarize spatial features based on their importance to question rather than considering each region equally. 
We are motivated by these approaches and propose to extend spatio-temporal reasoning to dialogues. 
However, rather than fixing on processing spatial inputs then learning temporal inputs, we note that in some cases, e.g. extended videos over a long period, it is more practical to first identify the relevant video segments before pinpointing the specific subjects of interest.
Considering questions in a dialogue setting, it is appropriate to assume the questions are relevant to varying temporal locations of the video rather than just a small fixed segment. 
We, thus, propose to explore a bidirectional vision-language reasoning approach to fully exploit both spatial and temporal-level features through two reasoning directions.

Our approach includes two parallel networks to learn relevant visual signals from the input video based on the language signals from user utterances.
Each network projects the language-based features to a three-dimensional tensor which is then used to independently learn video signals following a reasoning direction either as \textit{spatial$\rightarrow$temporal} or \textit{temporal$\rightarrow$spatial}. 
The output from each network is dynamically combined by importance scores computed based on language and visual features.
The weighted output is recurrently used as input to the reasoning modules to allow the models to progressively derive relevant video signals over multiple steps.
Intuitively, \textit{spatial$\rightarrow$temporal} reasoning is more appropriate for human queries related to specific entities or for input video involving many objects. \textit{temporal$\rightarrow$spatial} reasoning is more suitable for human queries about a particular video segment or for videos of extensive lengths. 


We name our proposed approach Bidirectional Spatio-Temporal Learning (\emph{BiST}), with the following contributions:
(1) Rather than exploiting temporal-level information only, our approach equally emphasizes both spatial and temporal features of videos for higher-resolution queries of visual cues.
(2) To tackle the diverse queried information from conversational queries, we propose a bidirectional strategy, denoted \textit{spatial$\leftrightarrow$temporal}, to enable comprehensive information diffusion between the two visual feature spaces. 
(3) Our models achieve competitive performance on the ``AVSD" (Audio-Visual Scene Aware Dialogues) benchmark from the 7$^{th}$ Dialogue System Technology Challenge (DSTC7) \cite{alamri2018audio, alamri2019audiovisual}. We adapt our models to a video QA task ``TGIF-QA" \cite{jang2017tgif} and achieve significant performance gains. 
(4) We conduct a comprehensive ablation and qualitative analysis and demonstrate the efficacy of our bidirectional reasoning approach. 

\section{Related Work}
\noindent Our work is related to two research topics: video-grounded dialogues and spatio-temporal learning.

\noindent \textbf{Video-grounded Dialogues}.
Following recent efforts that combine NLP and Computer Vision research \cite{antol2015vqa, xu2015show, goyal2017making}, video-grounded dialogues are extended from the two major research fields: video action recognition and detection \cite{simonyan2014two, yang2016multilayer, carreira2017quo} and dialogues/QA \cite{rajpurkar2016squad, budzianowski-etal-2018-multiwoz, gao2019neural}.
Approaches to video-grounded dialogues \cite{sanabria2019cmu, hori2019avsd, le-etal-2019-multimodal} typically use pretrained video models, such as 2D CNN models on video frames \cite{donahue2015long, feichtenhofer2016convolutional}, and 3D CNN models on video clips \cite{tran2015learning, carreira2017quo}, to extract visual features. 
However, these approaches mostly exploit the superficial information from the temporal dimension and neglect spatial-level signals. 
These approaches integrate spatial-level features simply through sum pooling with equal weights to obtain a global representation at the temporal level. 
They are, thus, not ideal for complex questions that investigate entity-level or spatial-level information \cite{jang2017tgif, alamri2019audiovisual}.
The dialogue setting exacerbates this limitation as it allows users to explore various aspects of the video contents, including both low-level (spatial) and high-level (temporal) information, over multiple dialogue turns. 
Our approach aims to address this challenge in video-grounded dialogues by retrieving fine-grained information from video through a bidirectional reasoning framework. 

\noindent \textbf{Spatio-temporal Learning}.
Most efforts in spatio-temporal learning focus on action recognition or detection tasks. 
\cite{yang2019step} proposes to progressively refine coarse-scale information through temporal extension and spatial displacement for action detection.
\cite{Li_2019_CVPR} uses a shared network of 2D CNNs over three orthogonal views of video to obtain spatial and temporal signals for action recognition. 
\cite{qiu2019learning} adopts a two-path network architecture that integrates global and local information of both temporal and spatial dimensions for video classification. 
Other research areas that investigate spatio-temporal learning include video captioning \cite{aafaq2019spatio}, video super-resolution \cite{li2019fast}, and video object segmentation \cite{xu2019spatiotemporal}. 
In general, spatio-temporal learning approaches aim to process higher-resolution information from complex videos that involve multiple objects in each video frame or motions over video segments \cite{yang2019step}. 
We are motivated by a similar reason observed in video-grounded dialogues and explore a vision-language bidirectional reasoning approach to obtain more fine-grained visual features.

\section{BiST Model}
The input includes a video $V$, dialogue history of $(t-1)$ turns (where $t$ is the current turn), each including a pair of (human utterance $H$, dialogue agent response $A$) $(H_1, A_1,..., H_{t-1}, A_{t-1})$, and current human utterance $H_t$. The output is a system response $A_t$ that can address current human utterance.
The input video can contain features in different modalities, including vision, audio, and text (such as video caption or subtitle). 
Without loss of generalization, we can denote each text input as a sequence of tokens, each represented by a unique token index from a vocabulary set $V$: dialogue history $X_{\mathrm{his}}$, user utterance $X_{\mathrm{que}}$, text input of video $X_{\mathrm{cap}}$ 
, and output response $Y$. 
We also denote $L_\mathrm{S}$ as the length of a sequence $S$. For instance, $L_\mathrm{que}$ is the length of $X_\mathrm{que}$.
\begin{figure*}[htbp]
	\centering
	\resizebox{1.0\textwidth}{!} {
	\includegraphics{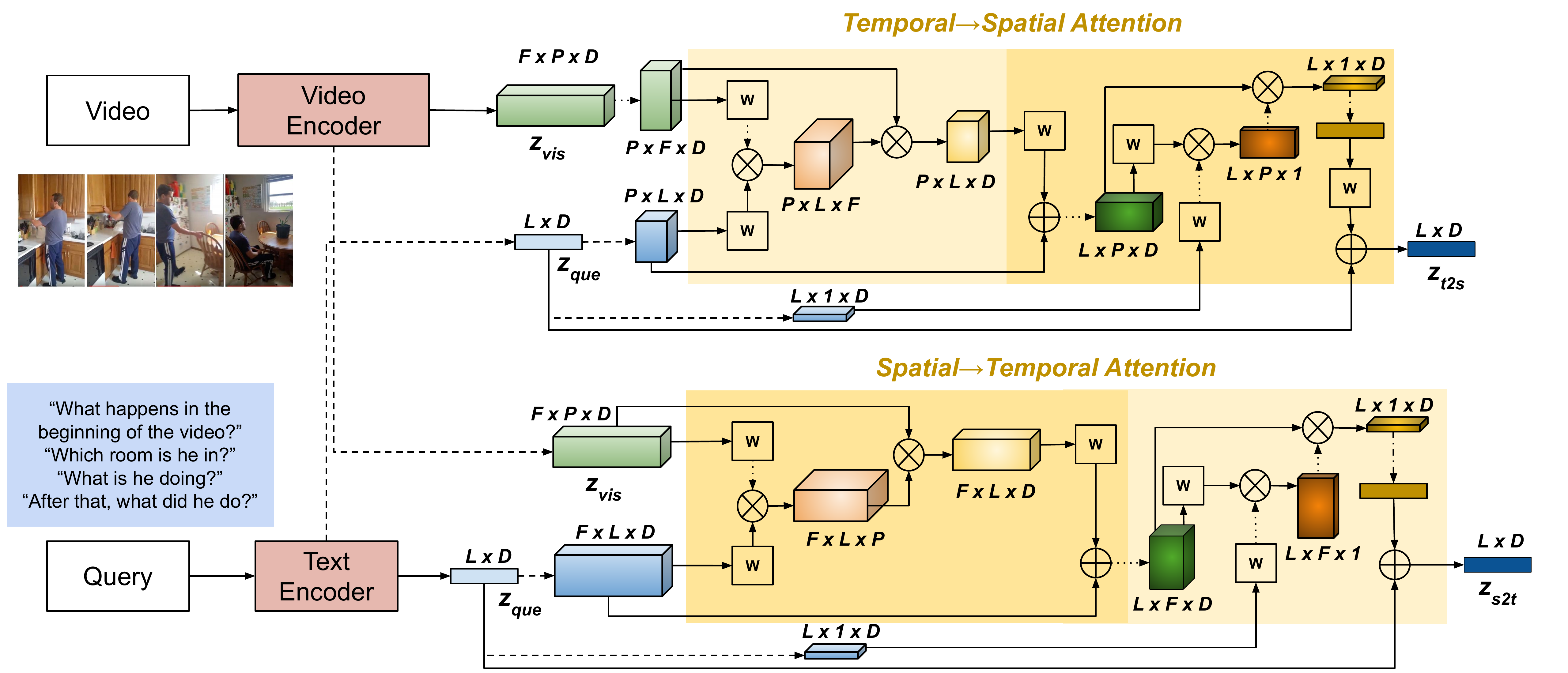}
	}
	\caption{Our bidirectional approach models the dependencies between text and vision in two reasoning directions: spatial$\rightarrow$temporal and temporal$\rightarrow$spatial.
	$\otimes$ and $\oplus$ denote dot-product operation and element-wise summation. 
	}
	\label{fig:bi_attn}
\end{figure*}

Our model is composed of four parts:
(1) The encoders encode text sequences and video inputs, including visual, audio, and text features, into continuous representations. 
For non-text features such as vision and sound, we follow previous work \cite{lei-etal-2018-tvqa, hori2019avsd} and assume access to pre-trained models.
(2) Several neural reasoning components learn dependencies between user utterances/queries and video features of multiple modalities.
For video visual features, we propose to learn dependencies at both spatial and temporal levels in two directions (see Figure \ref{fig:bi_attn}).
Specifically, we allow interaction between each token in user query and each spatial position or temporal step of the video. 
The outputs from spatial-based or temporal-based reasoning are sequentially incorporated in two directions, \textit{temporal$\rightarrow$spatial} and \textit{spatial$\rightarrow$temporal}.
The bidirectional strategy enables information being fused dynamically and captures complex dependencies between textual signals from dialogues and visual signals from videos. 
(3) The decoder passes encoded system responses over multiple attention steps, each of which integrates information from textual or video representations.
The decoder output is passed to a generator to generate tokens by an auto-regressive way. 
(4) The generator computes three distributions over the vocabulary set, one distribution as output from a linear transformation and the others based on pointer attention scores over positions of input sequences.
\subsection{Encoders}
\noindent \textbf{Text Encoder}. We use an encoder to embed text-based input $X$ into continuous representations $Z \in \mathbb{R}^{L_{X} \times d}$. 
$L_X$ is the length of sequence $X$ and $d$ is the embedding dimension. 
A text encoder includes a token-level embedding layer and a layer normalization \cite{ba2016layer}. 
The embedding layer includes a trainable matrix $E \in \mathbb{R}^{|V| \times d}$, with each row representing a token in the vocabulary set $V$ as a vector of dimension $d$.
We denote $E(X)$ as the embedding function that looks up the vector of each token in input sequence $X$: $Z_{\mathrm{emb}} = E(X) \in \mathbb{R}^{L_{X} \times d}$.
To incorporate the positional encoding layer, we adopt the approach from \cite{vaswani17attention} with each token position represented as a sine or cosine function. 
The output from positional encoding and token-level embedding is combined through element-wise summation and layer normalization.
The encoder outputs include representations for dialogue history $Z_{\mathrm{his}}$, user query $Z_{\mathrm{que}}$, video caption $Z_{\mathrm{cap}}$, and target response $Z_{\mathrm{res}}$.
For target response, during training, the sequence is shifted left by one position to allow prediction in the decoding step $i$ is auto-regressive on the previous positions $1,...,(i-1)$. We share the embedding matrix $E$ to encode all text sequences.

\noindent \textbf{Video Encoder}. 
We make use of a 3D-CNN video model to extract spatio-temporal visual features.
The dimensions of the resulting output depend on the configuration of sampling stride and clip length. 
We denote the output from a pretrained visual model as $Z^{\mathrm{pre}}_{\mathrm{vis}} \in \mathbb{R}^{F \times P \times d^{\mathrm{pre}}_{\mathrm{vis}}}$ where $F$ is the number of sampled video clips, $P$ is the spatial dimension from a 3D CNN layer, and $d^{\mathrm{pre}}_{\mathrm{vis}}$ is the feature dimension. 
We apply a linear layer with ReLU and layer normalization to reduce feature dimension to $d \ll d^{\mathrm{pre}}_{\mathrm{vis}}$.
For audio features, we follow similar procedure to obtain audio representation $Z_{\mathrm{aud}} \in \mathbb{R}^{F \times d}$.
We keep the pretrained visual and audio models fixed and directly use extracted features to our dialogue models.
\subsection{Bi-directional Reasoning}
\noindent We propose a bidirectional architecture whereby the text features are used to select relevant information in both spatial and temporal dimensions in two reasoning directions (See Figure \ref{fig:bi_attn}).

\noindent \textbf{Temporal$\rightarrow$spatial.} In one direction, the user query is used to select relevant information along temporal steps of each spatial region independently.
We first stack the encoded query features to $P$ spatial positions and denote the stacked features as $Z^{\mathrm{stack}}_{\mathrm{que}} \in \mathbb{R}^{P \times L_{\mathrm{que}} \times d}$.
For each spatial position, the model learns the dependencies between question and each of $F$ temporal steps through an attention mechanism as follows: 
\begin{align}
    Z^{(1)}_{t2s} &= Z^\mathsf{T}_{\mathrm{vis}}  W^{(1)}_{t2s} \in \mathbb{R}^{P \times F \times d_{\mathrm{att}}} \label{eq1}\\
    Z^{(2)}_{t2s} &= Z^{\mathrm{stack}}_{\mathrm{que}} W^{(2)}_{t2s} \in \mathbb{R}^{P \times L_{\mathrm{que}} \times d_{\mathrm{att}}} \label{eq2}
\end{align}
\begin{align}
    S^{(1)}_{t2s} &= \mathrm{Softmax}(Z^{(2)}_{t2s} {Z^{(1)}_{t2s}}^\mathsf{T} ) \in \mathbb{R}^{P \times L_{\mathrm{que}} \times F} \label{eq3}
\end{align}
where $d_\mathrm{att}$ is the dimension of the attention hidden layer, $W^{(1)}_{t2s} \in \mathbb{R}^{d \times d_{att}}$ and $W^{(2)}_{t2s} \in \mathbb{R}^{d \times d_{att}}$. The attention scores $S^{(1)}_{t2s}$ are used to obtain weighted sum along the temporal dimension of each spatial position of $Z_{\mathrm{vis}}$. The resulting tensor is passed through a linear transformation and ReLU layer. The output contains temporally attended visual features and are combined with language features through skip connection. We denote the output by vector $Z^{t}_{t2s}$.

\noindent From the temporally attended features, user query is used again to obtain dependencies along the spatial dimension. We use a similar attention network to model the interaction between each token in query and each temporally attended spatial region.
\begin{align}
    Z^{(3)}_{t2s} &=Z^{t}_{t2s} W^{(3)}_{t2s} \in \mathbb{R}^{L_{\mathrm{que}} \times P \times d_{\mathrm{att}}}\\
    Z^{(4)}_{t2s} &= Z_{\mathrm{que}}W^{(4)}_{t2s} \in \mathbb{R}^{L_{\mathrm{que}} \times d_{\mathrm{att}}} \label{eq11}\\
    S^{(2)}_{t2s} &= \mathrm{Softmax}(Z^{(3)}_{t2s} {Z^{(4)}_{t2s}}^\mathsf{T}) \in \mathbb{R}^{L_{\mathrm{que}} \times P} \label{eq6}
\end{align}
where $W^{(3)}_{t2s} \in \mathbb{R}^{d \times d_{\mathrm{att}}}$ and $W^{(4)}_{t2s} \in \mathbb{R}^{d \times d_{\mathrm{att}}}$.
The attention scores $S^{(2)}_{t2s}$ is used to obtain the weighted sum of all spatial positions from $Z^{t}_{t2s}$.
The output is temporal-to-spatially attended visual features and is incorporated into language features through skip connection. We denote the resulting output as $Z_{t2s}$.

\noindent \textbf{Spatial$\rightarrow$temporal.} 
In this reasoning direction, similar neural operations are used to compute spatially attended features followed by temporally attended features.
The main difference from the other reasoning direction is that we stacked the query features to $F$ temporal steps to obtain $Z^\mathrm{stack}_\mathrm{que} \in \mathbb{R}^{F \times L_\mathrm{que} \times d}$. Other network components, including two attention layers, are as described in Equation \ref{eq1} to \ref{eq6}. The final output is denoted as $Z_{s2t}$.

\noindent Previous approaches in video-based NLP tasks \cite{yu2016video,jang2017tgif,hori2019avsd} focus on the interaction between global representations of questions and temporal-level representations of videos. 
This strategy potentially loses critical information on spatial variations in video frames.
Our approach does not only emphasize both spatial and temporal feature spaces but also
allows neural models to diffuse information from these feature spaces in two different ways. 
As we can consider spatial information as local signals and temporal information as global signals, our approach enables global-to-local and local-to-global diffusion of visual cues in video. 
This approach is similar to \cite{qiu2019learning} in which local and global visual signals are learned and diffused iteratively. 
However, different from this approach, our approach focuses on language-vision reasoning for more accurate visual information queries. 

\noindent \textbf{Multimodal Reasoning.}
In addition to language-vision reasoning, our models also consider learning of other information dependencies between queries and audio inputs or textual video inputs. 
\begin{itemize}
\item Language$\rightarrow$Audio Reasoning. We adopt similar neural operations from language-vision reasoning. The difference is that we directly use the query features without stacking the features into Equation \ref{eq1} to \ref{eq3}. 
The resulting output of text-audio reasoning is denoted as $Z_\mathrm{q2a}$ which contains query-guided temporally attended features of $Z_\mathrm{aud}$.
\item Language$\rightarrow$Language Reasoning.
This reasoning module focuses on the unimodal dependencies between user query and video caption (if the caption is available). As the caption can contain useful information about the video content, we apply the dot-product attention mechanism similarly as with audio features to obtain $Z_{\mathrm{q2c}}$.
\end{itemize}

\noindent \textbf{Multimodal Fusioning}. Given the attended features, we combine them to obtained query-guided video representation, incorporating information from all modalities. 
We denote the concatenated representation in the following: $$Z_{\mathrm{q2vid}}=[Z_\mathrm{que}; Z_{t2s}; Z_{s2t}, Z_{\mathrm{q2a}}, Z_\mathrm{q2c}] \in \mathbb{R}^{L_{\mathrm{que}} \times 5d}$$ where $;$ is the concatenation operation.
The features are combined through an importance score matrix: 
$$S_{\mathrm{vid}} = \mathrm{Softmax}(Z_{\mathrm{q2vid}} W_{\mathrm{q2vid}}) \in \mathbb{R}^{L_{\mathrm{que}} \times 4}$$
where $W_{\mathrm{q2vid}} \in \mathbb{R}^{5d \times 4}$. 
The scores from $S_{\mathrm{vid}}$ are used to obtain the weighted sum of component video modalities, resulting in a fusion vector from multiple modalities.
We denote the resulting output $Z_{\mathrm{vid}}$.
Compared to previous work such as \cite{hori2019avsd, le-etal-2019-multimodal} which generally treat all modalities equally, our multimodal features are fused in a question-dependent manner. Potentially, our approach can avoid noisy or unnecessary signals, e.g. audio features not needed for questions only concerning visual contents. 
\subsection{Response Decoder}
\noindent The decoder aims to decode system responses in an auto-regressive manner. 
During inference, a special token $\langle$sos$\rangle$ is fed to the decoder. The output token is then concatenated to this special token as input to the decoder again to decode the second token. This repeats until reaching a limit of decoding rounds or when the special token $\langle$eos$\rangle$ is predicted. 
We apply a similar decoding architecture as \cite{le-etal-2019-multimodal}.
The decoder includes three attention layers to incorporate contextual cues from textual components to the output token representations. 
The first layer is a self-attention to learn dependencies among the current tokens. 
Intuitively, this helps to shape a more semantically structured sequence.
The second and third attention steps are used to capture contextual information from dialogue history and current user query to make the responses coherently connected to the whole dialogue context. 
To incorporate contextual cues from video components, our decoder is slightly different from \cite{le-etal-2019-multimodal}.
Instead of sequentially going through multiple attention layers, we only need one layer on the fused features $Z_{\mathrm{vid}}$.
This is more memory efficient since it only requires a single attention operation. 
It also does not depend on the design decision of the ordering of attention layers. 
At decoding step $j$, we denote the decoder output as $Z_\mathrm{dec} \in \mathbb{R}^{j \times d}$. 


\subsection{Pointer Generator}
\noindent Given the output from the decoder, the generator network is used to materialize responses in natural language.
A linear transformation is used to obtain distribution over the vocabulary set $V$.
\begin{align*}
P_{\mathrm{vocab}} = \mathrm{Softmax}(Z_{\mathrm{dec}} W_{\mathrm{vocab}}) \in \mathbb{R}^{j \times |V|}
\end{align*}
where $W_{\mathrm{vocab}} \in \mathbb{R}^{d \times |V|}$. 
We share the weights between $W_{\mathrm{vocab}}$ and $E$ as the semantics between source sequences and target responses are similar.

\noindent To strengthen the model generation capability, we adopt pointer networks \cite{nips2015pointer} to emphasize tokens from source sequences, i.e. user queries and video captions. 
We denote $\mathrm{Ptr}(Z_1, Z_2)$ as the pointer network operation i.e. each token in $Z_2$ is ``pointed'' to all tokens in $Z_1$ through a learnable probability distribution.
The resulting probability distribution is aggregated by all tokens in $Z_1$ to obtain $\mathrm{Ptr}(Z_1, Z_2) \in L_{Z_2} \times |V|$.
The final output distribution, denoted $P_\mathrm{out} \in \mathbb{R}^{j \times |V|}$, is the weighted sum of three distributions: $P_\mathrm{vocab}$, $\mathrm{Ptr}(Z_{\mathrm{que}}, Z_{\mathrm{dec}})$, and $\mathrm{Ptr}(Z_{\mathrm{cap}}, Z_{\mathrm{dec}})$. 
The weights for this fusion are learned via a linear transformation with softmax:
$\alpha = \mathrm{Softmax}(Z_{\mathrm{gen}}W_{\mathrm{gen}}) \in \mathbb{R}^{L_{\mathrm{res}} \times 3}$
where $Z_{\mathrm{gen}} = [Z_{\mathrm{res}}; Z_{\mathrm{dec}}; Z^{\mathrm{exp}}_{\mathrm{que}}; Z^{\mathrm{exp}}_{\mathrm{cap}}] \in \mathbb{R}^{j \times 4d}$, $W_{gen} \in \mathbb{R}^{4d \times 3}$, and $Z^{\mathrm{exp}}_{\mathrm{que}}$ and $Z^{\mathrm{exp}}_{\mathrm{cap}}$ are the stacked tensors of caption and user queries to $j$ dimensions.

\noindent \textbf{Optimization.}
During training, we learn all model parameters by minimizing the generation loss: 
\begin{align*}
\mathcal{L} = \sum_{j=0}^{L_Y} -\log(P_\mathrm{out}(y_j)).
\end{align*}

\section{Experiments}
\subsection{Experimental Setups}
\noindent \textbf{Datasets.} We use the AVSD benchmark from DSTC7 \cite{alamri2018audio, alamri2019audiovisual} which contains dialogues grounded on the Charades videos \cite{sigurdsson2016hollywood}.
In addition, we adapt our models to the video QA benchmark TGIF-QA \cite{jang2017tgif}. 
(See Table \ref{tab:datasets} for a summary of the two datasets). 
To extract visual and audio features, we used 3D-CNN ResNext-101 \cite{xie2017aggregated} pretrained on Kinetics \cite{hara2018can} to obtain spatio-temporal visual features and VGGish pretrained on YouTube videos \cite{hershey2017cnn} to extract (temporal) audio features. 
We sample video clips to extract visual features with a window size of 16 frames, and stride of 16 and 4 in AVSD and TGIF-QA respectively. 
In TGIF-QA experiments, we also extract visual features from pretrained ResNet-152 \cite{he2016deep} for a fair comparison with existing work.
In AVSD experiments, we make use of the video summary as the video-dependent text input $X_{\mathrm{cap}}$. 

\begin{table}[htbp]
	\centering
	\resizebox{1.0\columnwidth}{!} {
	\begin{tabular}{cllll}
    \hline
    \multicolumn{1}{l}{Benchmark}             & \textbf{\#}     & \multicolumn{1}{c}{\textbf{Train}} & \multicolumn{1}{c}{\textbf{Val.}} & \multicolumn{1}{c}{\textbf{Test}} \\ \hline
    \multirow{3}{*}{\textbf{AVSD}}   & Dialogs & 7,659                              & 1,787                             & 1,710                             \\ \ 
                                     & Turns   & 153,180                            & 35,740                            & 13,490                            \\ 
                                     & Words   & 1,450,754                          & 339,006                           & 110,252                           \\ \hline
    \multirow{4}{*}{\textbf{TGIF-QA}} & Count QA   & 24,159                             & 2,684                             & 3,554                             \\ 
                                     & Action QA  & 18,428                             & 2,047                             & 2,274                             \\ 
                                     & Trans. QA  & 47,434                             & 5,270                             & 6,232                             \\ 
                                     & Frame QA   & 35,453                             & 3,939                             & 13,691                            \\ \hline
    \end{tabular}
    }
	\caption{Summary of DSTC7 AVSD and TGIF-QA benchmark.
	The TGIF-QA contains 4 different tasks: (1) Count: open-ended QA which counts the number of repetitions of an action. (2) Action: multi-choice (MC) QA about a certain action occurring a fixed number of times. (3) Transition: MC QA about the temporal variation of video. (4) Frame: open-ended QA which can be answered from one video frame.
	}
	\label{tab:datasets}
\end{table}


\noindent \textbf{Training Procedure.}
We adopt the Adam optimizer \cite{kingma2014adam} and the learning rate strategy from \cite{vaswani17attention}. 
We set the learning rate \textit{warm-up} steps equivalent to 5 epochs and train models up to $50$ epochs.
We select the best models based on the average loss per epoch in the validation set. 
We initialize all model parameters with uniform distribution \cite{glorot2010understanding}. 
During training, we adopt the auxiliary auto-encoder loss function from \cite{le-etal-2019-multimodal}.
We adopt Transformer attention \cite{vaswani17attention} in our models and select the following hyper-parameters: 
$d=d_{\mathrm{att}}=128$, $N_{\mathrm{att}}=N_{\mathrm{dec}}=3$, and $h_{\mathrm{att}}=8$ where $N_{\mathrm{att}}$ and $N_{\mathrm{dec}}$ are the number of Transformer blocks in multimodal reasoning and decoder networks and $h_{\mathrm{att}}$ is the number of attention heads.
We tuned other hyper-parameters following grid-search over the validation set. 
In AVSD experiments, we train our models by applying label smoothing \cite{szegedy2016rethinking} on the target system responses $Y$. We adopt a beam search technique with a beam size $5$. 

\begin{table*}[htbp]
\centering
\resizebox{1.0\textwidth}{!} {
\begin{tabular}{lllllllllll}
\hline
\multicolumn{1}{c}{\textbf{Model}} & \textbf{$Z_\mathrm{vis}$} & \textbf{$Z_\mathrm{aud}$} & \textbf{$Z_\mathrm{cap}$} & \textbf{BLEU1} & \textbf{BLEU2} & \textbf{BLEU3} & \multicolumn{1}{c}{\textbf{BLEU4}} & \multicolumn{1}{c}{\textbf{METEOR}} & \textbf{ROUGE-L} & \textbf{CIDEr} \\
\hline
Baseline \cite{hori2019avsd}                           & I3D             & -               & -               & 0.621          & 0.480          & 0.379          & 0.305                              & 0.217                               & 0.481            & 0.733          \\
MTN \cite{le-etal-2019-multimodal}                               & I3D             & -               & -               & 0.654          & 0.521          & 0.420          & 0.343                              & 0.247                               & 0.520            & 0.936          \\
MTN \cite{le-etal-2019-multimodal}                               & ResNext         & -               & -               & 0.688          & 0.550          & 0.444          & 0.363                              & 0.260                               & 0.541            & 0.985          \\
\textbf{BiST}                               & ResNext         & -               & -               & \textbf{0.711} & \textbf{0.578} & \textbf{0.475} & \textbf{0.394}                     & \textbf{0.261}                      & \textbf{0.550}   & \textbf{1.050} \\
\hline
Video Sum. \cite{sanabria2019cmu}                       & ResNext         & -               & \checkmark               & 0.718          & 0.584          & 0.478          & 0.394                              & 0.267                               & 0.563            & 1.094          \\
Video Sum.+How2 \cite{sanabria2019cmu}                       & ResNext         & -               & \checkmark               & 0.723          & 0.586          & 0.476          & 0.387                              & 0.266                               & 0.564            & 1.087          \\
MTN \cite{le-etal-2019-multimodal}                               & I3D             & -               & \checkmark               & 0.715          & 0.581          & 0.476          & 0.392                              & 0.269                               & 0.559            & 1.066          \\
MTN \cite{le-etal-2019-multimodal}                               & ResNext         & -               & \checkmark               & 0.731          & 0.597          & 0.490          & 0.406                              & 0.271                               & 0.564            & 1.127          \\
\textbf{BiST}                               & ResNext         & -               & \checkmark               & \textbf{0.754} & \textbf{0.622} & \textbf{0.515} & \textbf{0.430}                     & \textbf{0.284}                      & \textbf{0.584}   & \textbf{1.190} \\
\hline
Baseline \cite{hori2019avsd}                           & I3D             & VGGish          & -               & 0.626          & 0.485          & 0.383          & 0.309                              & 0.215                               & 0.487            & 0.746          \\
Baseline+GRU+HierAttn. \cite{le2019end}                           & I3D             & VGGish          & -               & 0.631          & 0.491          & 0.390          & 0.315                              & 0.239                               & 0.509            & 0.848          \\
FA+HRED \cite{nguyen2018film}                           & I3D             & VGGish          & -               & 0.648          & 0.505          & 0.399          & 0.323                              & 0.231                               & 0.510            & 0.843          \\
Student-Teacher \cite{hori2019joint}                   & I3D             & VGGish          & -               & 0.675          & 0.543          & 0.446          & 0.371                              & 0.248                               & 0.527            & 0.966          \\
MTN \cite{le-etal-2019-multimodal}                               & I3D             & VGGish          & -               & 0.692          & 0.556          & 0.450          & 0.368                              & 0.259                               & 0.537            & 0.964          \\
MTN \cite{le-etal-2019-multimodal}                               & ResNext         & VGGish          & -               & 0.688          & 0.554          & 0.452          & 0.372                              & 0.251                               & 0.531            & 0.950          \\
\textbf{BiST}                               & ResNext         & VGGish          & -               &  \textbf{0.715}              &  \textbf{0.560}              &   \textbf{0.477}             &     \textbf{0.390}                               &     \textbf{0.259}                                &    \textbf{0.552}              &    \textbf{1.030}            \\
\hline
Baseline+GRU+HierAttn. \cite{le2019end}                           & I3D             & VGGish          & \checkmark               & 0.633          & 0.490          & 0.386          & 0.310                              & 0.242                               & 0.515            & 0.856          \\
FA+HRED \cite{nguyen2018film}                           & I3D             & VGGish          & \checkmark               & 0.695          & 0.553          & 0.444          & 0.360                              & 0.249                               & 0.544            & 0.997          \\
Student-Teacher \cite{hori2019joint}                   & I3D             & VGGish          & \checkmark               & 0.727          & 0.593          & 0.488          & 0.405                              & 0.273                               & 0.566            & 1.118          \\
MTN \cite{le-etal-2019-multimodal}                               & I3D             & VGGish          & \checkmark               & 0.731          & 0.597          & 0.494          & 0.410                              & 0.274                               & 0.569            & 1.129          \\
MTN \cite{le-etal-2019-multimodal}                               & ResNext         & VGGish          & \checkmark               & 0.735          & 0.600          & 0.498          & 0.413                              & 0.275                               & 0.571            & 1.137          \\
\textbf{BiST}                               & ResNext         & VGGish          & \checkmark               &  \textbf{0.755}              &     \textbf{0.619}           &   \textbf{0.510}             &    \textbf{0.429}                                &       \textbf{0.284}                              &   \textbf{0.581}               &    \textbf{1.192}     \\
\hline
\end{tabular}
}
\caption{Evaluation results on the test split of the AVSD benchmark. The results are presented in 4 settings by video feature components: (1) visual-only, (2) visual and text, (3) visual and audio, and (4) visual, audio, and text.}
\label{tab:avsd_results}
\end{table*}

\subsection{Modifications for Video QA}
\label{subsec:videoqa_modify}
\noindent In many Video QA benchmarks such as TGIF-QA \cite{jang2017tgif}, the tasks are retrieval-based (e.g. output a single score for each output candidate) rather than generation-based as in many dialogue tasks. 
Following \cite{fan2019heterogeneous}, we first concatenate the question with each candidate answer individually and treat this as $Z_{\mathrm{que}}$ to our models.
As there is no target response to be decoded, we adapt our models to this setting by using a trainable vector $z_j \in \mathbb{R}^{d}$ to represent a candidate response $R_j$, replacing $Z_{\mathrm{res}} \in \mathbb{R}^{j \times d}$ in a dialogue, as input to the decoder. 
The output, denoted $Z_{j,\mathrm{dec}} \in \mathbb{R}^{d}$,
is passed to a linear transformation layer to obtain a score $s_{j,\mathrm{out}}=Z_{j,\mathrm{dec}} W_\mathrm{out} \in \mathbb{R}$ where $W_\mathrm{out} \in \mathbb{R}^{d \times 1}$.
In this setting, we remove the language$\rightarrow$language and language$\rightarrow$audio reasoning modules.  
The loss function is the summed pairwise hinge loss \cite{jang2017tgif} between scores of positive answer $s^p_\mathrm{out}$ and each negative answer $s^n_{j,\mathrm{out}}$. $\mathcal{L} = \sum_{j=1}^{K} max(0, m - (s^p_\mathrm{out} - s^n_{j,\mathrm{out}}))$ where $K$ is the total number of candidate answers and $m$ is a hyper-parameter used as a margin between positive and negative answers.

\noindent \textbf{Training.}
Multiple-choice tasks, including \emph{Action} and \emph{Transition}, are trained following the pairwise loss with $K=5$ and $m=1$. 
\emph{Count} task is trained with similar approach but as a regression problem with a single output score $s_\mathrm{out}$. The loss function is measured as mean square error between output $s_{\mathrm{out}}$ and label $y$.
The open-ended \emph{Frame} task is trained as a generation task, similarly to the dialogue response generation task, with a single-token output. 
We use the the vector $z \in \mathbb{R}^{d}$ as input to the decoder. 
The generator includes a single linear layer with $W_\mathrm{out} \in \mathbb{R}^{d \times |v|}$. We do not apply pointer network in this case as the output is only a single-token response. 


\subsection{Results}
\noindent \textbf{AVSD Results.} We report the objective scores, including BLEU \cite{papineni2002bleu}, METEOR \cite{banerjee2005meteor}, ROUGE-L \cite{lin2004rouge}, and CIDEr \cite{vedantam2015cider}. 
These metrics, which formulate lexical overlaps between generated and ground-truth dialogue responses, are borrowed from language generation tasks such as machine translation and captioning.
We compare our generated responses with 6 reference responses. Major baseline models are: 
(1) \textit{Baseline} \cite{alamri2018audio, hori2019avsd} consists of LSTM-based encoder-encoder with attention layers between user queries and temporal-level visual and audio features.
(2) \textit{Baseline+GRU+HierAttn.} \cite{le2019end} extends (1) through GRU and question-guided self-attention and caption attention. 
(3) \textit{FA+HRED} \cite{nguyen2018film} adopts FiLM neural blocks for language-vision dependency learning. 
(4) \textit{Video Summarization} \cite{sanabria2019cmu} reformulates the task as a video summarization task and enhances the models with transfer learning from a large-scale summarization benchmark. 
(5) \textit{Student-Teacher} \cite{hori2019joint} adopts dual network architecture in which a student network is trained to mimic a teacher network trained with additional video-dependent text input. 
(6) \textit{MTN} \cite{le-etal-2019-multimodal} fuses temporal features of different modalities sequentially through a Transformer decoder architecture.   
(7) \textit{FGA} \cite{schwartz2019factor} consists of attention networks between all pairs of modalities and the models aggregate attention scores along edges of an attention graph. 

\begin{table*}[htbp]
\centering
\resizebox{0.65\textwidth}{!} {
\begin{tabular}{llllll}
\hline
\multicolumn{1}{c}{\textbf{Model}} & \textbf{$Z_\mathrm{vis}$} & \multicolumn{1}{c}{\textbf{\begin{tabular}[c]{@{}c@{}}Count \\ ( Loss)\end{tabular}}} & \multicolumn{1}{c}{\textbf{\begin{tabular}[c]{@{}c@{}}Action \\ (Acc)\end{tabular}}} & \multicolumn{1}{c}{\textbf{\begin{tabular}[c]{@{}c@{}}Trans. \\ (Acc)\end{tabular}}} & \multicolumn{1}{c}{\textbf{\begin{tabular}[c]{@{}c@{}}Frame \\ (Acc)\end{tabular}}} \\
\hline
VIS (aggr) \cite{ren2015exploring}                         & R              & 5.09                                                                                               & 0.468                                                                                               & 0.569                                                                                               & 0.346                                                                                       \\
VIS (avg) \cite{ren2015exploring}                         & R              & 4.80                                                                                               & 0.488                                                                                               & 0.348                                                                                               & 0.350                                                                                       \\
MCB (aggr) \cite{fukui-etal-2016-multimodal}                        & R              & 5.17                                                                                               & 0.589                                                                                               & 0.243                                                                                               & 0.257                                                                                       \\
MCB (avg) \cite{fukui-etal-2016-multimodal}                         & R              & 5.54                                                                                               & 0.291                                                                                               & 0.330                                                                                               & 0.155                                                                                       \\
Yu \emph{et al.} \cite{yu2017end}                          & R              & 5.13                                                                                               & 0.561                                                                                               & 0.640                                                                                               & 0.396                                                                                       \\
ST-VQA (s) \cite{jang2017tgif}                        & R+C          & 4.28                                                                                               & 0.573                                                                                               & 0.637                                                                                               & 0.455                                                                                       \\
ST-VQA (t) \cite{jang2017tgif}                        & R+C          & 4.40                                                                                               & 0.608                                                                                               & 0.671                                                                                               & 0.493                                                                                       \\
ST-VQA (st) \cite{jang2017tgif}                       & R+C          & 4.56                                                                                               & 0.570                                                                                               & 0.596                                                                                               & 0.478                                                                                       \\
Co-Mem \cite{gao2018motion}                            & R+F      & 4.10                                                                                               & 0.682                                                                                               & 0.743                                                                                               & 0.515                                                                                       \\
PSAC \cite{li2019beyond}                              & R              & 4.27                                                                                               & 0.704                                                                                               & 0.769                                                                                               & 0.515                                                                                       \\
HME  \cite{fan2019heterogeneous}                              & R+C          & 4.02                                                                                               & 0.739                                                                                               & 0.778                                                                                               & 0.538                                                                                       \\
STA  \cite{gao2019structured}                              & R          & 4.25                                                                                               & 0.723                                                                                               & 0.790                                                                                               & 0.566                                                                                       \\
CRN+MAC  \cite{le2019learning}                          & R              & 4.23                                                                                               & 0.713                                                                                               & 0.787                                                                                               & 0.592                                                                                       \\
MQL \cite{leimulti}                                   & V                                    & -                     & -                     & -                         & 0.598                  \\
QueST \cite{jiangdivide}                                & R                                    & 4.19                  & 0.759                 & 0.810                     & 0.597                  \\
HGA \cite{jiangreasoning}                                   & R+C                                  & 4.09                  & 0.754                 & 0.810                     & 0.551                  \\
GCN  \cite{huang2020location}                                  & R+C                                  & 3.95                  & 0.743                 & 0.811                     & 0.563                  \\
HCRN  \cite{le2020hierarchical}                                 & R+RX                                 & 3.82                  & 0.750                 & 0.814                     & 0.559                  \\  
\hline 
\textbf{BiST}                               & R              &   2.40                            & 0.839                                                                                                    &    0.817                                                                                                 &  0.630          \\
\textbf{BiST}                               & RX             & \textbf{2.19}                                                                                               & \textbf{0.847}                                                                                               &   \textbf{0.819}                                                                                                  & \textbf{0.648}         \\\hline                                                                             
\end{tabular}
}
\caption{Evaluation results on the test split of the TGIF-QA benchmark. Visual features are: R(ResNet), C(C3D), F(FlowCNN), RX(ResNext).}
\vspace{-0.1in}
\label{tab:tgif_results}
\end{table*}

\noindent In Table \ref{tab:avsd_results}, we present the scores by different combinations of features, including vision $Z_\mathrm{vis}$, audio $Z_\mathrm{aud}$, and text $Z_\mathrm{cap}$. 
In all settings, our models outperform the existing approaches. 
The performance of our models in the visual-only setting shows the performance gain coming from our bidirectional language-vision reasoning approach.
We also observe a performance boost whenever the text feature from video is considered. 
When we add the audio features, however, the performance gain is not significant.
This reveals a potential future extension in our work to better combine visual and audio feature representations.
FGA \cite{schwartz2019factor} reports the CIDEr score of $0.806$ in the visual-only setting. 
Compared to FGA, our performance gain indicates the efficacy of learning fine-grained dependencies between query and visual features at both spatial and temporal levels to select relevant information from video.

\noindent \textbf{TGIF-QA Results.} We give the L2 loss for \emph{Count} task and accuracy for the other three QA tasks (See Appendix \ref{app:tgifqa_baselines} for description of baseline models).
From Table \ref{tab:tgif_results}, our model outperforms existing approaches across all QA tasks, using either frame-level (appearance) feature, ResNet, or sequence-level feature, ResNext. 
Our models perform better with ResNext as we expect sequence-level feature is more consistent than frame-level feature. 
Experiments on this benchmark show clearer performance gain of our bidirectional language-vision reasoning approach as the performance is not affected by errors of generation components as in the AVSD experiments.
By focusing on learning high-resolution dependencies from spatio-temporal features, our models can fully exploit contextual cues and select better answers for video QA tasks. 
\begin{figure*}[h]
	\centering
	\resizebox{1.0\textwidth}{!} {
	\includegraphics{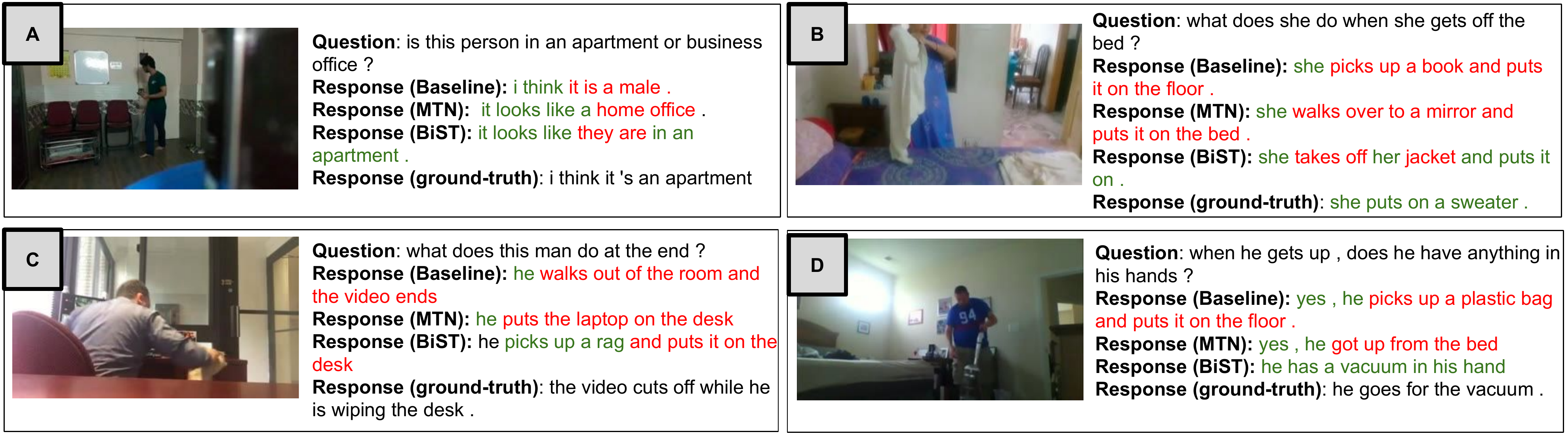}
	}
	\caption{Comparison of dialogue response outputs of BiST against the baseline models. Parts of the outputs that match and do not match the ground truth are highlighted in green and red respectively.}
	\label{fig:bist_examples}
\end{figure*}

\noindent \textbf{Impacts of Spatio-temporal Learning.} 
We consider model variants based on the spatio-temporal dynamics and report the results in Table \ref{tab:ablation}.
We noted that when using a single reasoning direction, the model with \textit{temporal$\rightarrow$spatial} performs better than one with the reverse reasoning direction.
This observation is different from prior approaches of spatio-temporal learning such as \cite{jang2017tgif} which are limited to the reasoning order \textit{spatial$\rightarrow$temporal}.
This can be explained as the videos in the AVSD benchmark are typically longer than other QA benchmarks.
It is practical to focus on temporal locations in frame sequences first before selecting spatial regions in individual frames.
In addition, dialogue queries are positioned in a multi-turn setting whereby each turn is relevant to different video segments as the dialogue evolves. 
Potentially, this observation indicates an important difference of video-grounded dialogues compared to video QA. 
Secondly, we also observe that our model performance improves when we use both reasoning directions rather than only one of them.
Our motivation for this approach is similar to \cite{schuster1997bidirectional} who proposes a bidirectional strategy to process sequences in both forward and backward directions.
Similarly, our approach exploits visual information through a bidirectional information diffusion strategy that can interpret information from both spatial or temporal aspects based on language input. 
Finally, we observe that using spatio-temporal features is better than only using one of them, demonstrating the importance of information in both dimensions.
To obtain $Z_\mathrm{vis}$ for spatial-only or temporal-only features, the spatio-temporal features are passed through an average pooling operation along the temporal or spatial dimensions respectively.

\begin{table}[htbp]
\centering
\resizebox{1.0\columnwidth}{!} {
\begin{tabular}{ccllll}
\hline
\textbf{t2s}    & \textbf{s2t}   & \multicolumn{1}{c}{\textbf{BLEU4}} & \multicolumn{1}{c}{\textbf{METEOR}} & \multicolumn{1}{c}{\textbf{ROUGE-L}} & \multicolumn{1}{c}{\textbf{CIDEr}} \\ \hline
  \checkmark              &    \checkmark                          & \textbf{0.430}                     & \textbf{0.284}                      & \textbf{0.584}                       & \textbf{1.190}                              \\
 \checkmark               &                             & 0.422                              & 0.281                               & 0.581                                & 1.183                              \\
                &    \checkmark                        & 0.420                              & 0.282                               & 0.579                                & 1.177                              \\
\multicolumn{2}{c}{t only}               & 0.419                              & 0.278                               & 0.573                                & 1.156                              \\
\multicolumn{2}{c}{s only}               & 0.418                              & 0.276                               & 0.570                                & 1.150                              \\
\hline
\end{tabular}
}
\caption{Ablation analysis on the AVSD benchmark
with variants of BiST by spatio-temporal dynamics.
}
\label{tab:ablation}
\end{table}

\noindent \textbf{Ablation Analysis.}
We conduct experiments with model variants of different hyper-parameter settings. Specifically, we vary the the number of attention rounds $N_\mathrm{att}$ and attention heads $h_\mathrm{att}$. 
From Table \ref{tab:model_depth}, we noted the contribution of the multi-round architecture to language-vision reasoning as the performance improves with larger reasoning steps, i.e. up to three attention rounds. 
However, we observe that as we increase to more than 3 reasoning steps, the model performance only improves slightly. 
We also note that using a multi-head attention mechanism is suitable for tasks dealing with information-intensive media such as video and dialogues.
The multi-head structure enables feature projection to multiple subspaces and capture complex language-vision dependencies. 
\begin{table}[htbp]
\centering
\resizebox{1.0\columnwidth}{!} {
\begin{tabular}{ccllll}
\hline
\textbf{$N$} & \textbf{$h_\mathrm{att}$} & \multicolumn{1}{c}{\textbf{BLEU4}} & \multicolumn{1}{c}{\textbf{METEOR}} & \multicolumn{1}{c}{\textbf{ROUGE-L}} & \multicolumn{1}{c}{\textbf{CIDEr}} \\ \hline
3-3        & 8               & \textbf{0.430}                     & \textbf{0.284}                      & \textbf{0.584}                       & 1.190                              \\
            1-1        & 8               & 0.418                              & 0.280                               & 0.574                                & 1.171                              \\
                2-2        & 8               & 0.422                              & 0.278                               & 0.576                                & 1.171                              \\
3-3        & 1               & 0.414                              & 0.278                               & 0.580                                & 1.173                              \\
3-3        & 2               & 0.418                              & 0.280                               & 0.579                                & 1.174                              \\
3-3        & 4               & 0.428                              & 0.280                               & \textbf{0.584}                       & \textbf{1.195}                     \\ \hline
\end{tabular}
}
\caption{ 
Performance of model variants by  $N=N_\mathrm{att}=N_\mathrm{dec}$, and $h_\mathrm{att}$ on the AVSD benchmark
}
\label{tab:model_depth}
\end{table}

\noindent \textbf{Qualitative Analysis}. In Figure \ref{fig:bist_examples}, we present some example outputs. 
We note that the predicted dialogue responses of BiST models are closer to the ground-truth responses. 
Particularly for complex questions that query specific segments (example B, C, D), and/or specific spatial locations (Example D), our approach can generally produce better responses. 
Another observation is that for ambiguous examples such as Example C (where the visual appearance is not clear  to differentiate ``apartment'' and ``business office''), our model can return the correct answer. Potentially this can be explained by the extracted signals from spatial-level feature representations.
Finally, we note that there are still some errors that make the output sentences partially wrong, such as mismatching subjects (example A), wrong entities (Example B), or wrong actions (Example C). 
For detailed qualitative analysis, please refer to Appendix \ref{app:qual_analysis}.
\section{Conclusion}
\noindent We proposed BiST, a novel deep neural network approach for video-grounded dialogues and video QA, which exploits the complex visual nuances of videos through a bidirectional reasoning framework in both spatial and temporal dimensions.
Our experimental results show that BiST can extract relevant, high-resolution visual cues from videos and generate quality dialogue responses/answers. 



\clearpage
\break

\bibliography{anthology,emnlp2020}
\bibliographystyle{acl_natbib}

\clearpage
\break

\appendix

\section{TGIF-QA Baselines}
\label{app:tgifqa_baselines}
In TGIF-QA experiments, we compare our models with the following baselines:
(1) \textit{VIS} \cite{ren2015exploring} and 
(2) \textit{MCB} \cite{fukui-etal-2016-multimodal} are two image-based VQA baselines which were adapted to TGIF-QA by \cite{jang2017tgif}.
(3) Yu \emph{et al.} \cite{yu2017end} uses a high-level concept word detector and the detected words are used for semantic reasoning.
(4) \textit{ST-VQA} \cite{jang2017tgif}  integrates temporal and spatial features by first pre-training temporal part and then finetuning the spatial part.
(5) \textit{Co-Mem} \cite{gao2018motion} includes a co-memory mechanism on two video streams based on motion and appearance features.
(6) \textit{PSAC} \cite{li2019beyond} uses multi-head attention layers to exploit the dependencies between text and temporal variation of video.
(7) \textit{HME} \cite{fan2019heterogeneous} is a memory network with read and write operations to update global context representations. 
(8) \textit{STA} \cite{gao2019structured} divides video into $N$ segments and uses temporal attention modules on each segment independently. 
(9) \textit{CRN+MAC} \cite{le2019learning} is a clip-based reasoning framework by aggregating frame-level features into clips through temporal attention.
(10) \textit{MQL} \cite{leimulti} exploits the semantic relations among questions and proposes a multi-label prediction task. 
(11) \textit{QueST} \cite{jiangdivide} has two types of question embeddings: spatial and temporal embeddings based on attention guided by video features. 
(12) \textit{HGA} \cite{jiangreasoning} is a graph alignment network consisting of inter- and intra-modality edges to model the interaction between video and question. 
(13) \textit{GCN} \cite{huang2020location} is a similar approach with graph network but utilizes the video object-level features as node representations. 
(14) \textit{HCRN} \cite{le2020hierarchical} extends \cite{le2019learning} with a hierarchical relation network over temporal-level video features.
\begin{figure*}[h]
	\centering
	\resizebox{1.0\textwidth}{!} {
	\includegraphics{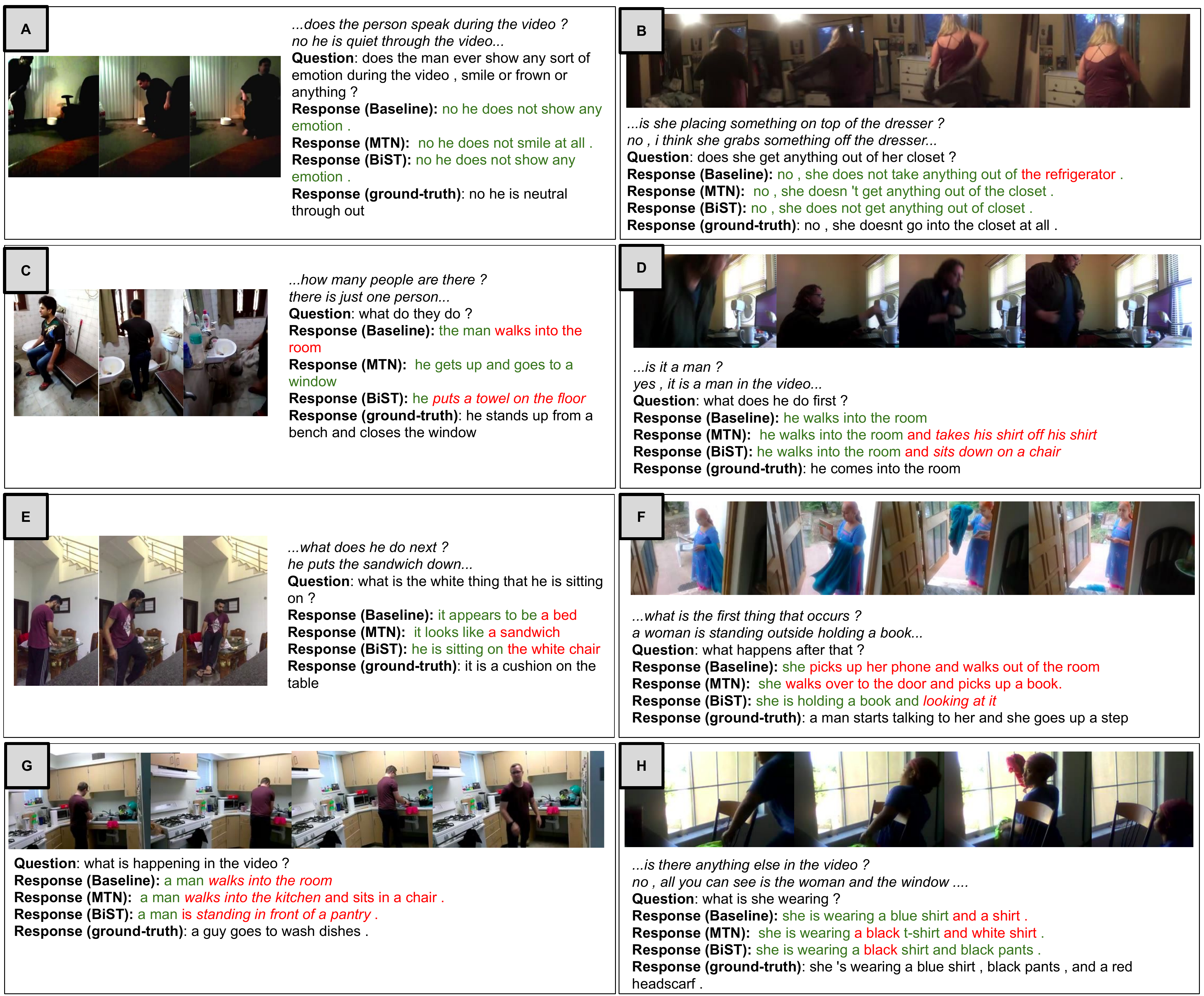}
	}
	\caption{Comparison of dialogue response outputs of BiST against the baseline models: \emph{Baseline} \cite{hori2019avsd} and \emph{MTN} \cite{le-etal-2019-multimodal}. Parts of the outputs that match and do not match the ground truth are highlighted in green and red respectively.}
	\label{fig:additional_bist_examples}
\end{figure*}

\section{Qualitative Analysis}
\label{app:qual_analysis}
We present additional example outputs in Figure \ref{fig:additional_bist_examples}. For each examples, we include the last dialogue turn from the dialogue history. 
In general, BiST can generate responses that better match the ground truth than the Baseline \cite{hori2019avsd} and MTN \cite{le-etal-2019-multimodal} (example A, B).
Furthermore, we analyze both negative and positive outputs and have the following observations: 
\begin{itemize}
    \item In cases where the videos contain more than one actions, our models can predict responses that describe multiple actions in their correct \emph{orders of appearance}. 
    For instance, in example D, even though our model response does not completely match the ground truth, it is still correctly explaining the sequence of actions, including first ``walking into the room'' and ``sits down on a chair'', matching the visual input from video. MTN response in the same example  can express multiple actions but fail to detect the second action before ``takes his shirt off''. A similar observation can be found in the example F. 
    \item In cases where the entities are hard to detect due to weak \emph{visual distinction}, BiST can materialize the correct entity in its responses, e.g. in example C, ``a towel'' was seen in the last sampled video frame. 
    Another example is example H where BiST detects both ``shirt'' and ``pants'' entities (even though their color attributes are not totally correct). 
    However, in example E, all models fail to identify the entity ``a cushion'', possibly because of the ambiguous and subdue visual features of this object in the video. 
    This displays an important challenge for more fine-grained information extraction in video-grounded dialogues. 
    \item We noted our model fails in the following complex cases. First, for case with \emph{ambiguous questions} such as example C, BiST emphasizes an action in the later part of the video (3$^{rd}$ sampled frame) rather than the early part of the video (1$^{st}$ and 2$^{nd}$ sampled frame). This error might be due to the implied temporal specification in the question.  
    Similarly, in example G, the ambiguous question results in generated responses of different action-level granularity from all models and some responses are partially correct.
    Secondly, in cases where the ground-truth answer involves \emph{unseen entity} (example F with the entity ``a man'' without any visual appearance but possibly detected by his voice in the audio input), our model fails to include this entity in the response. 
    A possible explanation for this example is that our model is not able to detect the entity based on audio input, i.e. ``a man talking''. This presents the retaining challenge to fully combine multiple modalities into natural language responses in dialogues. 
\end{itemize}

\end{document}